\documentclass[10pt,twocolumn,letterpaper]{article}

\usepackage{iccv}
\usepackage{times}
\usepackage{epsfig}
\usepackage{graphicx}
\usepackage{amsmath}
\usepackage{amssymb}
\usepackage[dvipsnames]{xcolor}
\usepackage{subcaption}
\usepackage{tabulary,multirow}
\newcommand{\bd}[1]{\textbf{#1}}
\newcolumntype{x}[1]{>{\centering\arraybackslash}p{#1pt}}
\newlength\savewidth\newcommand\shline{\noalign{\global\savewidth\arrayrulewidth
  \global\arrayrulewidth 1pt}\hline\noalign{\global\arrayrulewidth\savewidth}}
\newcommand{\tablestyle}[2]{\setlength{\tabcolsep}{#1}\renewcommand{\arraystretch}{#2}\centering\footnotesize}
\definecolor{demphcolor}{RGB}{124,124,124}
\newcommand{\demph}[1]{\textcolor{demphcolor}{#1}}

\usepackage[pagebackref=true,breaklinks=true,letterpaper=true,colorlinks,bookmarks=false]{hyperref}

\iccvfinalcopy 

\ificcvfinal\fi
\begin{document}

\title{ShapeMask: Learning to Segment Novel Objects by Refining Shape Priors}
\author{
Weicheng Kuo \thanks{Work done while at Google Brain}\\
Google Brain\\
UC Berkeley\\
{\tt\small weicheng@google.com}
\and
Anelia Angelova\\
Google Brain\\
{\tt\small anelia@google.com}
\and
Jitendra Malik\\
UC Berkeley\\
{\tt\small malik@eecs.berkeley.edu}
\and
Tsung-Yi Lin\\
Google Brain\\
{\tt\small tsungyi@google.com}
}

\maketitle

\begin{abstract}

Instance segmentation aims to detect and segment individual objects in a scene. Most existing methods rely on precise mask annotations of every category. However, it is difficult and costly to segment objects in novel categories because a large number of mask annotations is required. We introduce ShapeMask, which learns the intermediate concept of object shape to address the problem of generalization in instance segmentation to novel categories. ShapeMask starts with a bounding box detection and gradually refines it by first estimating the shape of the detected object through a collection of shape priors. Next, ShapeMask refines the coarse shape into an instance level mask by learning instance embeddings. The shape priors provide a strong cue for object-like prediction, and the instance embeddings model the instance specific appearance information. ShapeMask significantly outperforms the state-of-the-art by 6.4 and 3.8 AP when learning across categories, and obtains competitive performance in the fully supervised setting. It is also robust to inaccurate detections, decreased model capacity, and small training data. Moreover, it runs efficiently with 150ms inference time and trains within 11 hours on TPUs. With a larger backbone model, ShapeMask increases the gap with state-of-the-art to 9.4 and 6.2 AP across categories. Code will be released.

\end{abstract}
\vspace{-5mm}

\section{Introduction}

\begin{figure}[t]
\begin{center}
    \includegraphics[width=1.0\linewidth]{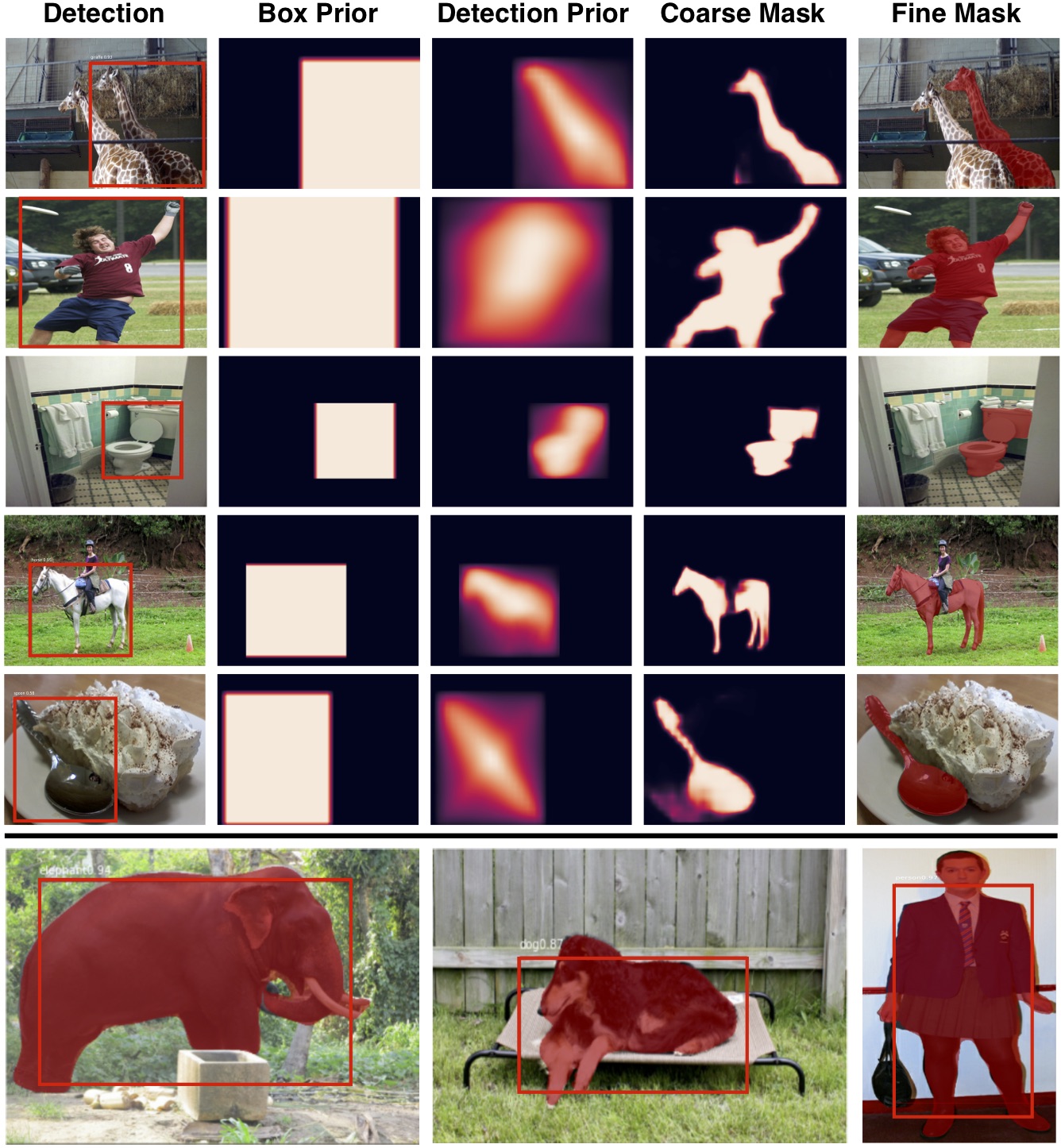}
\end{center}
\vspace{-5mm}
\caption{
ShapeMask instance segmentation is designed to learn the shape of objects by refining object shape priors. Starting from a bounding box (leftmost column), the shape is progressively refined in our algorithm until reaching the final mask (rightmost column). The bounding box is only needed to approximately localize the object of interest and is not required to be accurate (bottom row).
}
\vspace{-4mm}
\label{fig:illustration}
\end{figure}

Instance segmentation is the task of providing pixel-level classification of objects and identifying individual objects as separate entities. It is fundamental to applications such as autonomous driving or robot manipulation~\cite{cordts2016cityscapes,shu2018self}, since segmenting individual objects could help autonomous agents' planning and decision making. The community has made great headway on this task recently ~\cite{pinheiro2015learning,pinheiro2016learning,he2017mask,hariharan2014simultaneous,novotny2018semi,dai2016instance,kong2018recurrent,bai2017deep,liu2017sgn,hu2017learning,khoreva2017simple}. However, these approaches require precise \textit{pixelwise} supervision for \textit{every} category. The need for annotation limits instance segmentation to a small slice of visual world that we have dense annotations for. But how can instance segmentation generalize better to novel categories?

Most existing instance segmentation algorithms are either detection-based ~\cite{pinheiro2015learning,pinheiro2016learning,he2017mask,hariharan2014simultaneous,dai2016instance} and grouping-based \cite{liu2018affinity,arnab2017pixelwise,novotny2018semi,kong2018recurrent,bai2017deep,liu2017sgn}. To generalize to novel categories, detection-based approaches can use \textit{class-agnostic} training which treats all categories as one foreground category. For example, previous works perform figure-ground segmentation inside a box region without distinguishing object classes ~\cite{pinheiro2015learning,pinheiro2016learning}. Although class agnostic learning can be readily applied to novel categories, there still exists a clear gap compared to the fully supervised setup~\cite{hu2017learning,pinheiro2015learning}. On the other hand, the grouping-based approaches learn instance specific cues such as pixel affinity for grouping each instance. Although the grouping stage is inherently class-agnostic and suitable for novel categories, most algorithms still rely on semantic segmentation \cite{arnab2017pixelwise,liu2018affinity,bai2017deep} to provide class information, which requires pixelwise annotation of every class. Whether detection or grouping-based, generalization to novel categories remains an open challenge.

\begin{figure}
\begin{center}
   \includegraphics[width=0.95\linewidth]{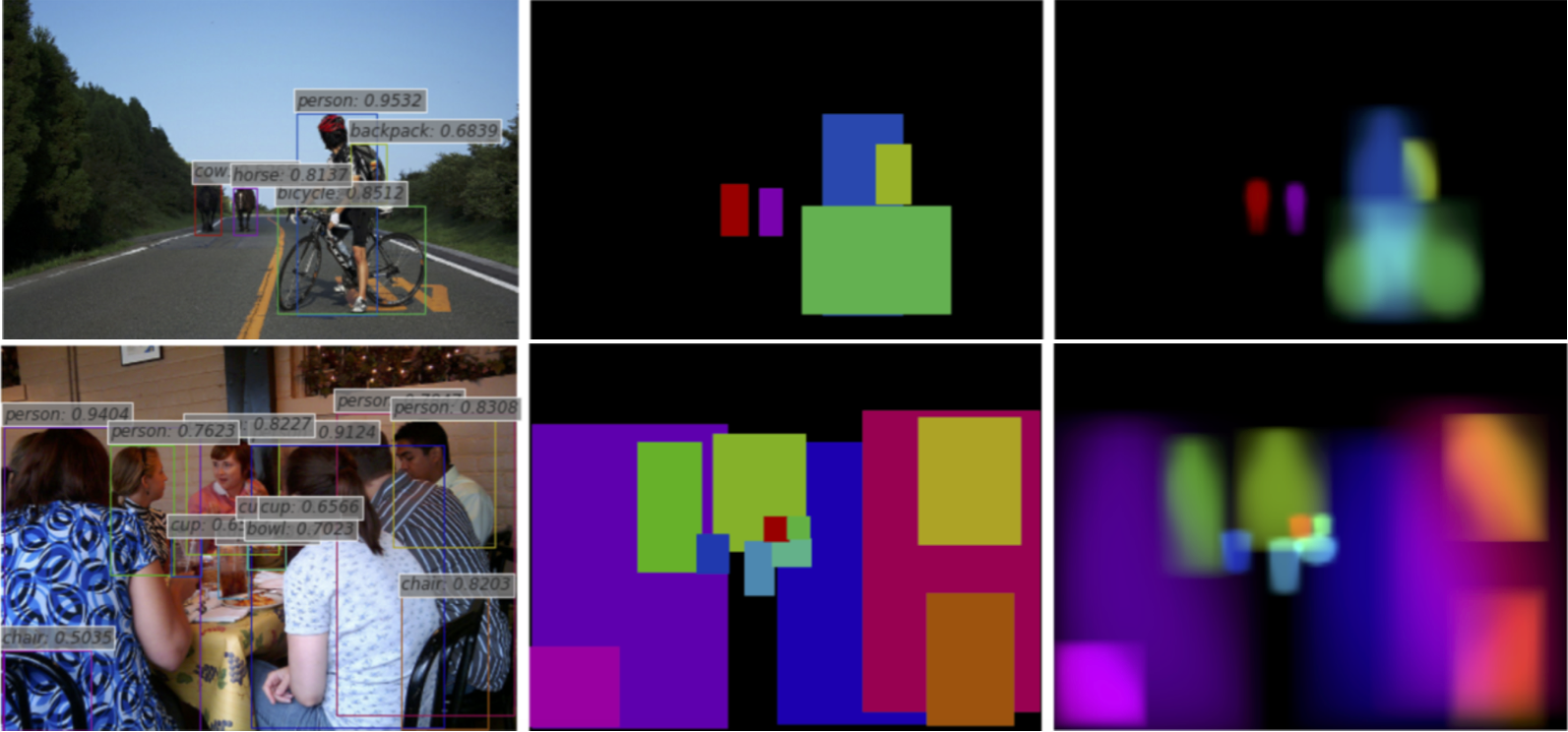}
\vspace{-5mm}
\end{center}
   \caption{Illustration of objects in uniform box priors vs. shape priors. Every row contains: (left) input image plus detections, (center) box priors, (right) shape priors. Shape priors represent objects with much richer details than boxes.}
 \vspace{-6mm}
\label{fig:detection-prior}
\end{figure}

We propose to improve generalization in instance segmentation by introducing intermediate representations \cite{lim2013sketch,sax2018mid,dave2019towards}, and instance-specific grouping-based learning \cite{pont2017multiscale,khoreva2017simple}. Consider Figure~\ref{fig:detection-prior}. Most detection-based approaches use boxes as the intermediate representation for objects (see middle column) which do not contain information of object pose and shape. On the contrary, shapes are more informative (see right column) and have been used by numerous algorithms to help object segmentation \cite{arnab2017pixelwise,yang2012layered,he2014exemplar,chen2015multi,weiss2013scalpel}. As the pixels of novel objects may appear very different, we hypothesize that shapes could be leveraged to improve generalization as well. Intuitively speaking, learning shapes may help because objects of different categories often share similar shapes, e.g., horse and zebra, orange and apple, fork and spoon. On the other hand, grouping-based learning causes the model to learn ``which pixels belong to the same object'' and may generalize well by learning appropriate appearance embeddings. For example, even if the model has never seen an orange before, it could still segment it by grouping the pixels with similar appearance.

Motivated by this observation, we propose a new instance segmentation algorithm ``ShapeMask'' to address the generalization problem. Figure~\ref{fig:illustration} illustrates how ShapeMask starts with a box detection, and gradually refines it into a fine mask by learning intermediate shapes. Given a detection, ShapeMask first represents it as a uniform box prior. Then ShapeMask finds the shape priors which best indicate the location, scale and rough shape of the object to fit the box (detection prior). Finally, ShapeMask decodes the coarse mask by a fully convolutional network and refines it by its own instance embedding. The idea behind refinement is similar to grouping approaches. To generalize to novel categories, we simply use class agnostic training for ShapeMask without the need of transfer learning. A natural by-product of learning shapes as soft priors is that ShapeMask can produce masks outside the detection box similar to \cite{hayder2017boundary} and unlike~\cite{he2017mask,dai2016instance} which apply feature cropping.

Furthermore we design ShapeMask to run seamlessly across hardware accelerators such as TPUs\cite{jouppi2017tpu,cloudtpu} and GPUs to maximize performance as follows: (1) perform simple cropping (no interpolation needed) instead of ROIAlign operation \cite{he2017mask}, (2) train with jittered ground truths instead of detections to avoid sorting or NMS at training time \cite{he2017mask}, (3) use one-stage detector RetinaNet \cite{lin2018focal} to enable efficient training. These design changes allow us to train 4x faster than state-of-the-art without compromising performance.

Experiments on COCO show that ShapeMask significantly outperforms the state-of-the-art transfer learning approach \cite{hu2017learning} in the cross-category setup. In fact, ShapeMask can outperform the state-of-the-art using only 1\% of the labeled data. We also qualitatively show that ShapeMask is able to segment many novel object classes in a robotics environment different from the COCO dataset. In the fully supervised instance segmentation setting, ShapeMask is competitive with state-of-the-art techniques while training multiple times faster and running at 150-200ms per image.

\section{Related Work}
\label{sec:related-works}
Most instance segmentation algorithms can be categorized into either \textit{detection-based} or \textit{grouping-based} approaches. 
The detection-based methods \cite{hariharan2014simultaneous,dai2016instance,hayder2017boundary,li2016fully,chen2017masklab,he2017mask,pinheiro2015learning,pinheiro2016learning} first detect the bounding box for each object instance and predict the segmentation mask inside the region cropped by the detected box. This approach has achieved state-of-the-art performance in instance segmentation datasets like COCO\cite{lin2014microsoft} and Cityscapes \cite{cordts2016cityscapes}. The grouping-based approaches \cite{kong2018recurrent,bai2017deep,Brabandere2017semantic,liu2018affinity,liu2017sgn,arnab2017pixelwise,kirillov2017instancecut} view the instance segmentation as a bottom-up grouping problem. They do not assign region of interest for each object instance. Instead, they produce pixelwise predictions of cues such as directional vectors \cite{liu2017sgn}, pairwise affinity \cite{liu2018affinity}, watershed energy \cite{bai2017deep}, and semantic classes, and then group object instances from the cues in the post-processing stage. In addition to grouping, some object segmentation works have simultaneously used \textit{shape priors} as unaries in probabilitic framework \cite{arnab2017pixelwise,yang2012layered,he2014exemplar}, augmented proposals \cite{chen2015multi}, or as top-down prior to help grouping \cite{weiss2013scalpel}. Classical instance segmentation works are mostly grouping-based and show minimal gap generalizing to unseen categories \cite{rother2004grabcut, pont2017multiscale}. For example, MCG \cite{pont2017multiscale} generates quality masks by normalized cut on the contour pyramid computed from low level cues. However, grouping-based approaches have not been shown to outperform detection-based methods on the challenging COCO dataset so far.

Recently, \cite{pham2018bayesian,zhou2018weakly,khoreva2017simple,hu2017learning} study instance segmentation algorithms that can generalize to categories without mask annotations. \cite{khoreva2017simple} leverages the idea that given a bounding box for target object, one can obtain pseudo mask label from a grouping-based segmentation algorithm like GrabCut \cite{rother2004grabcut}. \cite{pham2018bayesian} studies open-set instance segmentation by using a boundary detector followed by grouping, while \cite{zhou2018weakly} learns instance segmentation from image-level supervision by deep activation. Although effective, these approaches do not take advantage of \textit{existing} instance mask labels to achieve better performance.

In this paper, we focus on the \textit{partially supervised} instance segmentation problem~\cite{hu2017learning}, as opposed to the weakly-supervised setting \cite{khoreva2017simple,zhou2018weakly}. 
The main idea is to build a large scale instance segmentation model by leveraging large datasets with bounding box annotations e.g. \cite{krishna2017visual}, and smaller ones with detailed mask annotations e.g. \cite{lin2014microsoft}. More specifically, the setup is that only box labels (not mask labels) are available for a subset of categories at training time. The model is required to perform instance segmentation on these categories at test time. Mask$^{\mathit{X}}$ R-CNN\xspace \cite{hu2017learning} tackles the problem by learning to predict weights of mask segmentation branch from the box detection branch. This transfer learning approach shows significant improvement over class-agnostic training, but there still exists a clear gap with the fully supervised system.

\section{Method}

In the following sections, we discuss the set of modules that successively refine object box detections into accurate instance masks. 

\subsection{Shape Recognition}
\label{sec:shape_regcognition}

\begin{figure}[h]
    \begin{center}
      \includegraphics[width=0.9\linewidth]{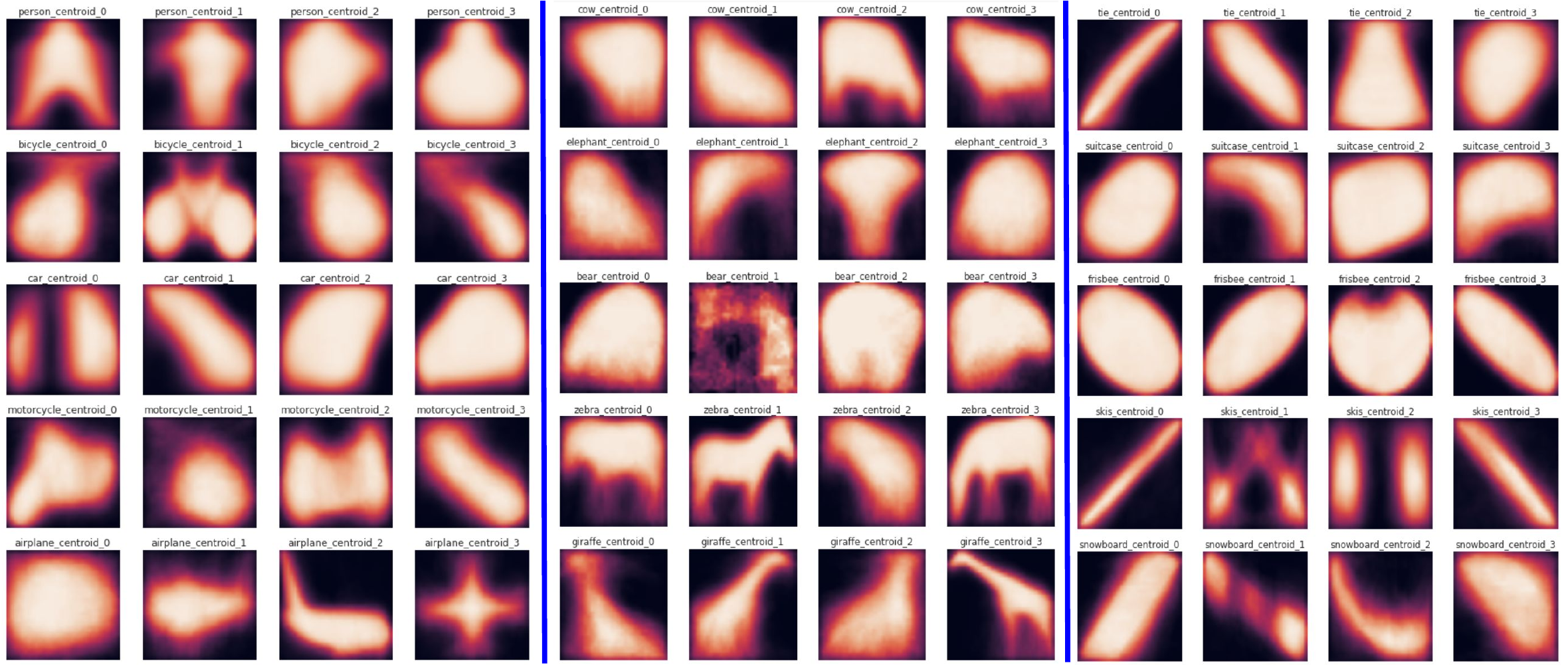}
    \end{center}
      \vspace{-4mm}
      \caption{Shape priors obtained by clustering mask labels in the training set. Each prior is a cluster centroid of an object category.}
      \vspace{-5mm}
    \label{fig:coco-centroids}
\end{figure}

\noindent\textbf{Shape priors:}
We obtain a set of shape bases from a collection of mask annotations in order to succinctly represent the canonical poses and shapes of each class. These bases are called ``shape priors''.
The intuition is that when the approximate shape is selected early on in the algorithm, the subsequent instance segmentation becomes much more informed than a box (see also Figure~\ref{fig:detection-prior}). In order to obtain shape priors, we run k-means to find $K$ centroids of all instance masks for \textit{each class} in the training set. We resize all mask annotations to a canonical size 32 x 32 before clustering. In the class specific setting, the total number of shape priors is $C \times K$, where $C$ is the number of classes (e.g. $K=20$). In the class agnostic setting, we group all classes as one and have $K$ shape priors in total (e.g., $K=100$).
We define the set of shape priors as $H=\{S_1,S_2,...,S_K\}$. Figure \ref{fig:coco-centroids} visualizes the selected shape priors per class for the COCO dataset. We can see the objects have diverse within- and between-class appearance. In class-agnostic setting, clustering yields similarly diverse shape priors.

\begin{figure}[t]
\begin{center}  
   \includegraphics[width=1.0\linewidth]{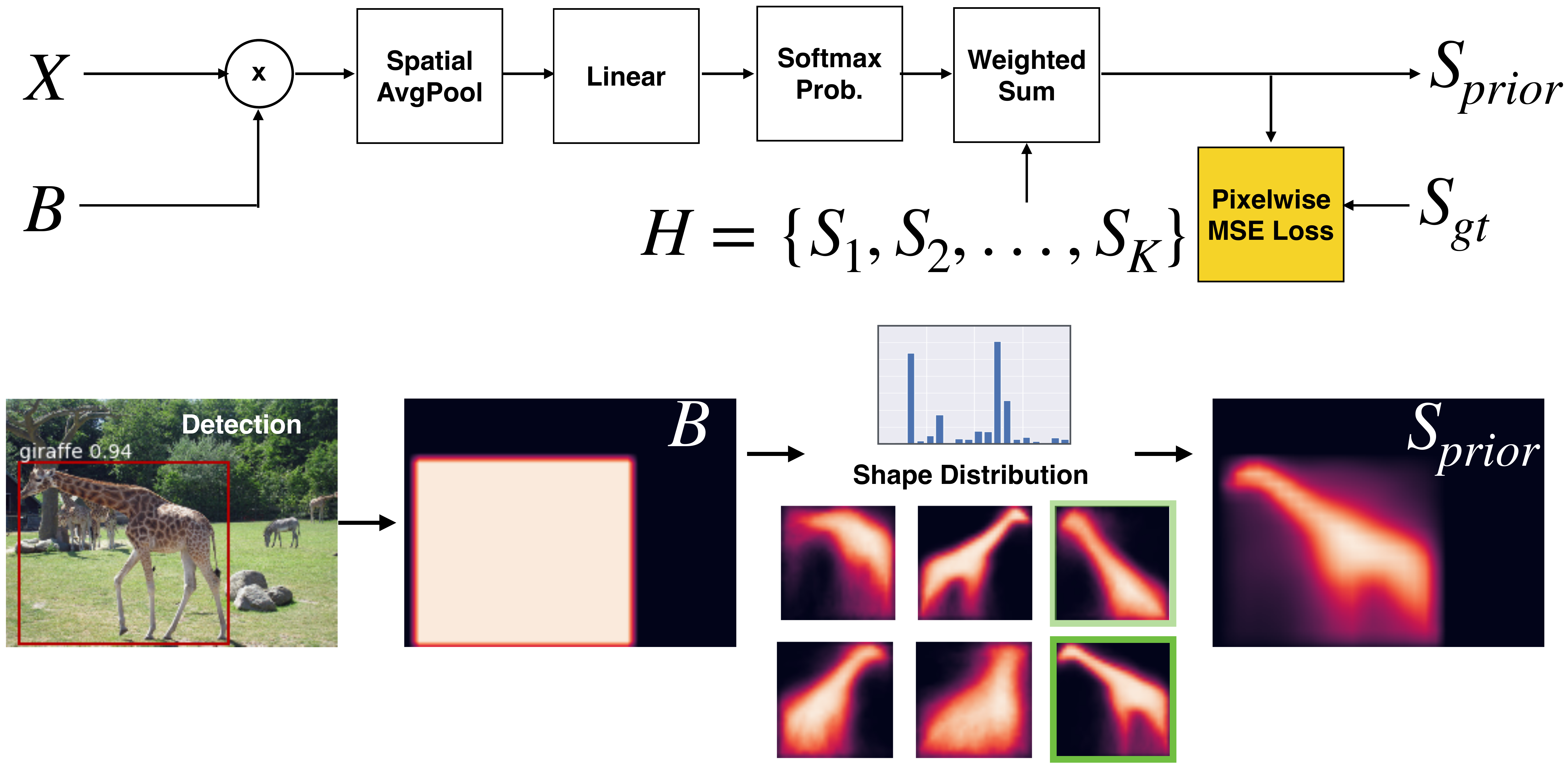}
\end{center}
\vspace{-4mm}
\caption{\textbf{Shape Estimation.} Given a box detection, we refine the box into an initial estimate of shape $S_{prior}$ by linearly combining prior shapes $S_1, S_2, ..., S_k$. Our model learns to predict the shape prior distribution to minimize reconstruction error. 
}
\vspace{-4mm}
\label{fig:shape-estimation}
\end{figure}

\noindent\textbf{Shape estimation:}
Starting with a box detection, we first represent it as a binary heatmap $B$, i.e. $b \in \{0, 1\}$, $\forall b \in B$. The purpose of this stage is to estimate a more informative detection prior $S_{prior}$ from $B$ (see Figure \ref{fig:shape-estimation}). To achieve this, we estimate the target object shape by selecting similar shape priors from the knowledge base $H$. Unlike existing methods \cite{chen2017masklab,he2017mask} which view shape prediction as a per-pixel classification problem, we learn to combine similar shapes from $H$ to form predictions. 

Figure \ref{fig:shape-estimation} illustrates the entire process. First, we pool features inside the bounding box $B$ on the feature map $X$, to obtain an embedding $x_{box}$ representing the object instance:
\begin{equation}
\label{eqn:shape_rec}
x_{box} = \frac{1}{|B|} \sum_{(i, j)\in B}X_{(i,j)}
\end{equation}
The instance shape embedding $x_{box}$ is then used to recognize similar shapes in the knowledge base $H$. The shape priors are the bases used to reconstruct the target object shape inside the bounding box. The predicted object shape $S$ is a weighted sum of shape priors $\{S_1, S_2, ..., S_K\}$, where the weights are predicted by applying a linear layer $\phi$ to $x_{box}$ followed by a softmax function to normalize weights over $K$, $w_k = softmax(\phi_k(x_{box}))$ 
\begin{equation}
\label{eqn:fuse_shape}
S = \sum_{k=1}^K  w_k S_k
\end{equation}
The predicted shape $S$ is then resized and fitted into the detection box $B$ to create a smooth heatmap, which we call ``detection prior'' $S_{prior}$ (as shown in Figure~\ref{fig:shape-estimation}). During training, we apply pixel-wise mean square error (MSE) loss on the detection prior $S_{prior}$ against the ground-truth mask $S_{gt}$ to learn the parameters in $\phi$ (See Equation \ref{eqn:shape_loss}).
\begin{equation}
\label{eqn:shape_loss}
L_{prior} = MSE(S_{prior}, S_{gt})
\end{equation}
The approach simplifies the instance segmentation problem by first solving the shape recognition problem. It incorporates the strong prior that the primitive object shapes only have a few modes. This regularizes the output space of the model and prevents it from predicting implausible shapes, e.g., broken pieces. By adding such structure to the model, we observe improved generalization to novel categories. We speculate this is because many novel objects share similar shapes with the labeled ones.

\subsection{Coarse Mask Prediction}
\label{sec:coarse_mask}

\begin{figure}
\includegraphics[width=1.0\linewidth]{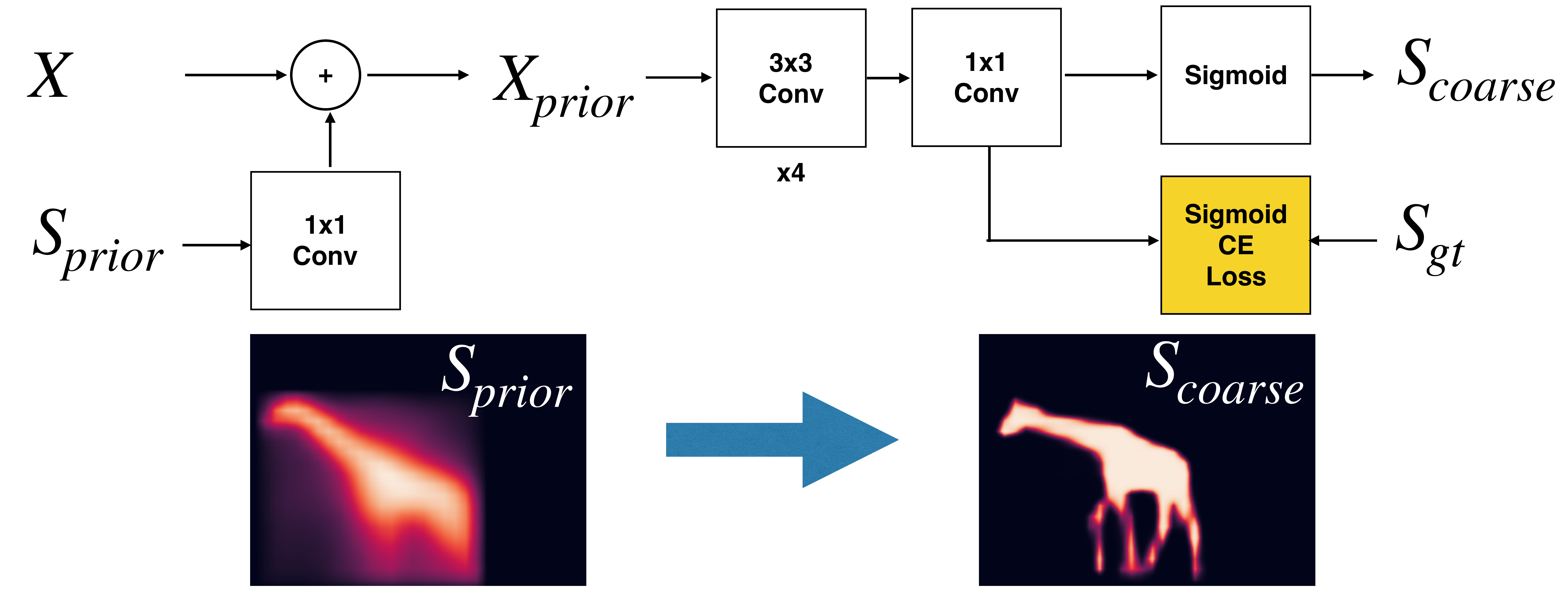}
\vspace{-5mm}
\caption{\textbf{Coarse Mask Prediction.} We fuse $S_{prior}$ with the image features $X$ to obtain prior conditioned features $X_{prior}$, from which we decode a coarse object shape $S_{coarse}$.
}
\vspace{-6mm}
\label{fig:coarse-mask}
\end{figure}

Given the detection prior $S_{prior}$ from the previous section, the goal of this stage is to obtain a coarse instance mask $S_{coarse}$ (Figure~\ref{fig:coarse-mask}). First, we use a function $g$ to embed $S_{prior}$ into the same feature dimension as image features $X$, where $g$ is a 1x1 convolution layer. Then we sum them into a prior conditioned feature map $X_{prior}$: 
\begin{equation}
\label{eqn:cond_prior}
X_{prior} = X + g(S_{prior})
\end{equation}
$X_{prior}$ now contains information from both image features and the detection prior which guides the network to predict object-like segmentation mask. A coarse instance mask $S_{coarse}$ is decoded by applying a function $f$ to $X_{prior}$, which consists of four convolution layers in our case: 
\begin{equation}
\label{eqn:coarse_mask}
S_{coarse} = f(X_{prior})
\end{equation}
This is similar to the mask decoder design in \cite{he2017mask}, but the difference is we use detection prior $S_{prior}$ to guide the decoding. Pixel-wise cross-entropy loss is applied to the predicted mask $S_{coarse}$ to learn the parameters in the mask decoder: 
\begin{equation}
\label{eqn:coarse_mask_loss}
L_{coarse} = CE(S_{coarse}, S_{gt})
\end{equation}


\subsection{Shape Refinement by Instance Embedding}
\label{sec:shape-refinement}
\begin{figure}
\includegraphics[width=1.0\linewidth]{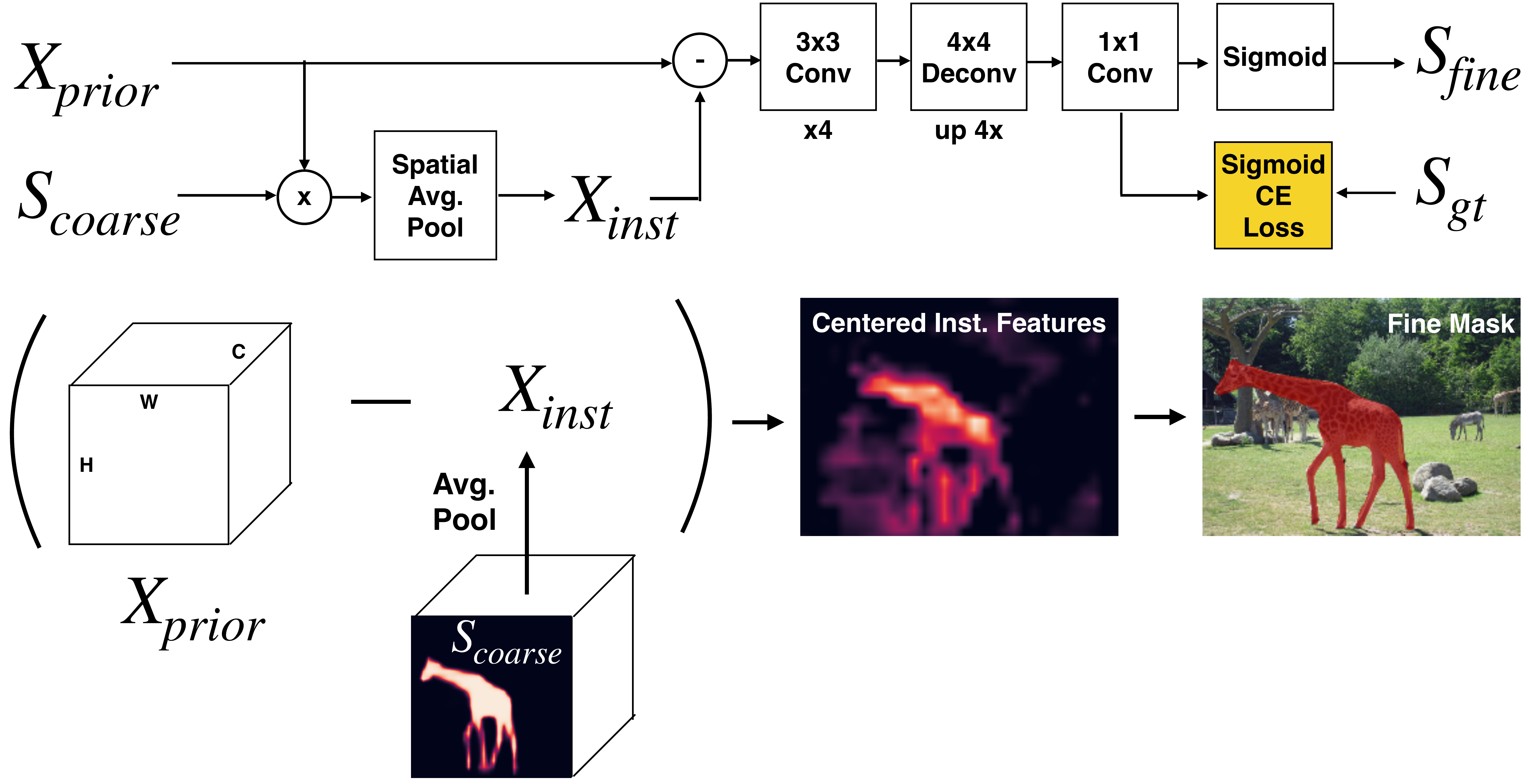}
\vspace{-5mm}
\caption{\textbf{Shape Refinement.} Starting from $X_{prior}$ and $S_{coarse}$, we first compute the instance embedding $X_{inst}$ by average pooling the features within $S_{coarse}$. Then we subtract $X_{inst}$ from $X_{prior}$ before decoding the final mask. We show the low-dimensional PCA projection of the ``Centered Instance features'' for the purpose of visualization.}
\vspace{-6mm}
\label{fig:shape-refinement}
\end{figure}

Although the coarse segmentation mask $S_{coarse}$ provides strong cues for possible object shapes, it does not leverage the instance-specific information encoded by the image features. As opposed to previous stages that aim to extract rough shape estimates, the goal of this stage is to refine $S_{coarse}$ into a detailed final mask $S_{fine}$ (Figure \ref{fig:shape-refinement}).

Similar to the instance shape embedding $x_{box}$ in Sec. \ref{sec:shape_regcognition}, we can pool the instance mask embedding by the refined shape prior to obtain more accurate instance representations $x_{mask}$.  Given the soft coarse mask $S_{coarse}$, we binarize it and compute the 1D instance embedding $x_{mask}$ of the target object by pooling features inside the coarse mask:
\begin{equation}
\label{eqn:pool_mask}
x_{mask} = \frac{1}{|S_{coarse}|} \sum_{(i, j)\in S_{coarse}} X_{prior(i,j)}
\end{equation}
We then center the image features $X_{prior}$ by subtracting the instance embedding $x_{mask}$ at all pixel locations: 
\begin{equation}
\label{eqn:subtract}
X_{inst(i,j)} = X_{prior(i,j)} - x_{mask}
\end{equation}
This operation can be viewed as conditioning the image features by the target instance. The idea is to encourage the model to learn simple, low-dimensional features to represent object instances. To obtain the fine mask $S_{fine}$, we add the mask decoding branch which has the same architecture as described in Section~\ref{sec:coarse_mask} with one additional upsampling layer to enhance the output resolution. Same as before, pixelwise cross-entropy loss is used to learn the fine mask $S_{fine}$ from the groundtruth mask $S_{gt}$ (see Equation \ref{eqn:fine_loss}).
\begin{equation}
\label{eqn:fine_loss}
L_{fine} = CE(S_{fine}, S_{gt})
\end{equation}
Note that the $S_{gt}$ here is of higher resolution than before due to the upsampling of $S_{fine}$.

\subsection{Generalization by Class Agnostic Learning}
To generalize to novel categories, we adopt class-agnostic learning in ShapeMask. We follow the setup in~\cite{hu2017learning}, the box branch outputs box detections with confidence scores for all classes and the mask branch predicts a foreground mask given a box without knowing the class. For generating shape priors ${S_1, S_2, ..., S_k}$, we combine instance masks from \textit{all} classes together and run k-means with a larger $K$ than the class-specific setting. This allows us more capacity to capture the diverse modes of shapes across all categories. At inference time, we treat any novel object as part of this one foreground category during shape estimation and mask prediction stages. The capability to generalize well across categories makes ShapeMask also a \textit{class-agnostic} algorithm, although its performance in the class-specific setting remains competitive among the best techniques.

\section{Implementation Details}
\label{sec:implementation}
\noindent\textbf{One-stage detector:} We adopt RetinaNet\footnote{https://github.com/tensorflow/tpu/tree/master/models/official/retinanet}~\cite{lin2018focal} to generate bounding box detections for ShapeMask. RetinaNet is a one-stage detector with a simple system design and competitive performance. We use an image size of 1024 x 1024 and multiscale training with image scale from 0.8 to 1.2. Note that other detection methods such as Faster R-CNN~\cite{ren2015faster} can also be used with ShapeMask.

\noindent\textbf{Training with jittered groundtruths:}
Unlike~\cite{he2017mask,chen2017masklab} which sample masks from the object proposals, we randomly sample a fixed number of groundtruth masks and their associated boxes per image for training (e.g. 8 in our case). This removes the need of object proposal stage and enables one-stage training for mask prediction. Additionally, the sampled groundtruth boxes are jittered by Gaussian noise to better mimic the imperfect boxes produced by the model during inference time. To be precise, the new box centers are $(x_c', y_c')=(x_c + \delta_x w, y_c + \delta_y h)$, and the new box sizes are $(w', h') = (e^{\delta_w} w, e^{\delta_h} h)$, where $(x,y,w,h)$, $(x',y',w',h')$ are the noiseless/jittered groundtruth boxes respectively, and $\delta$s are Gaussian noise $\sim N(\mu=0, \sigma=0.1)$. We represent these boxes by uniform box priors (see $B$ in Figure \ref{fig:shape-estimation}) for training the mask branch. Jittering is essential to help ShapeMask learn to be robust against noisy detections at test time.

\noindent\textbf{RoI features:} We use the feature pyramid~\cite{lin2017feature} with levels $P_3$ to $P_5$ to process RoI in different scales for scale normalization. Given a bounding box, we assign the box to feature level:
\begin{equation}
\label{eqn:level}
    k = m - \lfloor \log_2 \frac{L}{\max(box_h, box_w)}\rfloor,
\end{equation}
where $L$ is the image size (e.g., 1024) and $m$ is the highest feature level (e.g., 5). If $k$ is less than the minimum level (e.g. 3), the box is assigned to minimum level. At the assigned level, we take a $c \times c$ feature patch centered on the box. We choose $c = L /2^m$ to make sure the entire object instance always lies inside the patch by the feature pyramid design~\cite{lin2017feature}. The feature dimension is then reduced from 256 to 128 by 1x1 convolution. We apply ShapeMask algorithm on this feature patch $X$ to predict the instance mask (c.f. $X$ in Figure \ref{fig:shape-estimation}). Note that this does not require ``crop and resize'' operation~\cite{he2017mask,chen2017masklab}.

\begin{figure*}[t]
\begin{center}
  \includegraphics[width=0.8\textwidth]{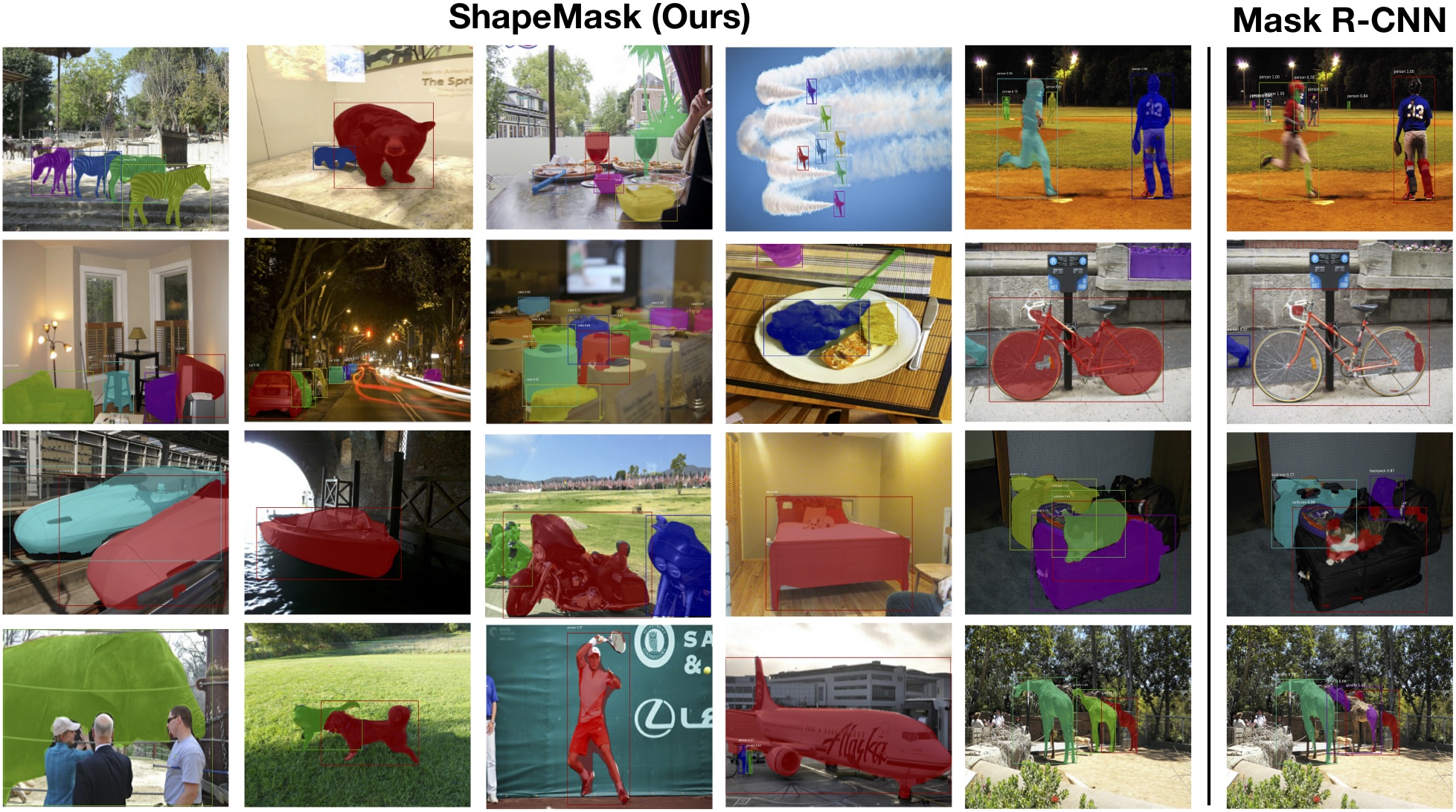}
\end{center}
\vspace{-5mm}
\caption{Visualization of ShapeMask on novel categories. For clarity, we only visualize the masks of novel categories. ShapeMask is able to segment many challenging objects well without seeing mask annotations in the same categories. It learns to predict object-like shapes for novel categories in many cases where Mask R-CNN does not (see rightmost column).
}
\vspace{-5mm}
\label{fig:vis-partial}
\end{figure*}

\begin{table*}[t]
\tablestyle{3.5pt}{1.1}
\begin{center}
\begin{tabular}{ll|x{22}x{22}x{22}x{22}x{22}x{22}|x{22}x{22}x{22}x{22}x{22}x{22}}
& & \multicolumn{6}{c|}{VOC $\rightarrow$ Non-VOC} & \multicolumn{6}{c}{Non-VOC $\rightarrow$ VOC} \\
backbone & method & AP & AP$_{50}$ & AP$_{75}$ & AP$_S$ & AP$_M$ & AP$_L$ & AP & AP$_{50}$ & AP$_{75}$ & AP$_S$ & AP$_M$ & AP$_L$ \\
\shline
& Mask R-CNN \cite{hu2017learning}&
18.5 & 34.8 & 18.1 & 11.3 & 23.4 & 21.7 & 24.7 & 43.5 & 24.9 & 11.4 & 25.7 & 35.1 \\
& Our Mask R-CNN &
21.9 & 39.6 & 21.9 & 16.1 & 29.7 & 24.6 & 27.2 & 39.6 & 27.0 & 16.4 & 31.8 & 35.4 \\
FPN & GrabCut Mask R-CNN \cite{hu2017learning}& 19.7 & 39.7 & 17.0 & 6.4 & 21.2 & 35.8 & 19.6 & 46.1 & 14.3 & 5.1 & 16.0 & 32.4 \\
& Mask$^{\mathit{X}}$ R-CNN\xspace \cite{hu2017learning}&
23.8 & 42.9 & 23.5 & 12.7 & 28.1 & 33.5 & 29.5 & 52.4 & 29.7 & 13.4 & 30.2 & 41.0 \\
& Oracle Mask R-CNN \cite{hu2017learning}
& \demph{34.4} & \demph{55.2} & \demph{36.3} & \demph{15.5} & \demph{39.0} & \demph{52.6} & \demph{39.1} & \demph{64.5} & \demph{41.4} & \demph{16.3} & \demph{38.1} & \demph{55.1} \\
& Our Oracle Mask R-CNN
& \demph{34.3} & \demph{54.7} & \demph{36.3} & \demph{18.6} & \demph{39.1} & \demph{47.9} & \demph{38.5} & \demph{64.4} & \demph{40.4} & \demph{18.9} & \demph{39.4} & \demph{51.4} \\
\hline
FPN & ShapeMask (ours) &
\bd{30.2} & \bd{49.3} & \bd{31.5} & \bd{16.1} & \bd{38.2} & \bd{38.4} & \bd{33.3} & \bd{56.9} & \bd{34.3} & \bd{17.1} & \bd{38.1} & \bd{45.4} \\
& Oracle ShapeMask (ours)
& \demph{35.0} & \demph{53.9} & \demph{37.5} & \demph{17.3} & \demph{41.0} & \demph{49.0} & \demph{40.9} & \demph{65.1} & \demph{43.4} & \demph{18.5} & \demph{41.9} & \demph{56.6} \\
\hline
NAS-FPN \cite{nasfpn} & ShapeMask (ours) &
\bd{33.2} & \bd{53.1} & \bd{35.0} & \bd{18.3} & \bd{40.2} & \bd{43.3} & \bd{35.7} & \bd{60.3} & \bd{36.6} & \bd{18.3} & \bd{40.5} & \bd{47.3} \\
& Oracle ShapeMask (ours)
& \demph{37.6} & \demph{57.7} & \demph{40.2} & \demph{20.1} & \demph{44.4} & \demph{51.1} & \demph{43.1} & \demph{67.9} & \demph{45.8} & \demph{20.1} & \demph{44.3} & \demph{57.8}  \\
\hline
\\
\end{tabular}
\end{center}
\vspace{-8mm}
\caption{Performance of ShapeMask (class-agnostic) on novel categories. At the top, VOC $\rightarrow$ Non-VOC means ``train on masks in VOC, test on masks in Non-VOC'', and vice versa. ShapeMask outperforms the state-of-the-art method Mask$^{\mathit{X}}$ R-CNN\xspace \cite{hu2017learning} by 6.4 AP on VOC to Non-VOC transfer, and 3.8 AP on Non-VOC to VOC transfer using the same ResNet backbone. ShapeMask has smaller gap with the oracle upper-bound than Mask$^{\mathit{X}}$ R-CNN\xspace. By using a stronger feature pyramid from \cite{nasfpn}, ShapeMask outperforms Mask$^{\mathit{X}}$ R-CNN\xspace by 9.4 and 6.2 AP. These results provide strong evidence that ShapeMask can better generalize to categories without mask annotation.}
\label{tab:partial_mask}
\vspace{-1.5em}
\end{table*}

\section{Experiments}

\paragraph{Experimental setup.} We report the performance of ShapeMask on the COCO dataset~\cite{lin2014microsoft}.
We adopt well established protocol in the literature for evaluation \cite{girshick2015fast, ren2015faster,he2017mask,dai2016instance,li2016fully,dai2016sensitive,chen2017masklab} by reporting standard COCO metrics AP, AP50, AP75, and AP for small/medium/large objects. Unless specified otherwise, mask AP is reported instead of box AP. We additionally compare the training and inference times, demonstrating ShapeMask is among the fastest at both inference and training.

\subsection{Generalization to Novel Categories}

We first demonstrate the state-of-the-art ability of ShapeMask to generalize across classes and datasets. 
Such generalization capability shows ShapeMask can work well on a larger part of the visual world than other approaches which require strong pixelwise labeling for every category. 

\label{sec:partial_sup}
Partially Supervised Instance Segmentation is the task of performing instance segmentation on a subset of categories for which no masks are provided during training. The model is trained on these categories with only box annotations, and on other categories with both box and mask annotations.
The experiments are set up following the previous work \cite{hu2017learning}. We split the COCO categories into VOC vs. Non-VOC. The VOC categories are those also present in PASCAL VOC \cite{everingham2010pascal}. At training time, our models have access to the bounding boxes of all categories, but the masks only come from either VOC or Non-VOC categories. The performance upper bounds are set by the oracle models that have access to masks from all categories. In this section, our training set is COCO train2017 and the comparison with other methods is done on val2017 Non-VOC/VOC categories following previous work \cite{hu2017learning}. 

\noindent \textbf{Main results:}
We achieve substantially better results than the state-of-the-art methods as shown in Table \ref{tab:partial_mask}. All benchmark experiments use ResNet-101 network with feature pyramid connections \cite{lin2017feature}. Using the same FPN backbone, ShapeMask outperforms the state-of-the-art method Mask$^{\mathit{X}}$ R-CNN\xspace \cite{hu2017learning} by 6.4 AP on VOC to Non-VOC transfer, and 3.8 AP on Non-VOC to VOC transfer. The gap relative to the oracle upper-bound is 4.8 and 7.6 AP for ShapeMask, compared to the 10.6 and 9.6 AP of Mask$^{\mathit{X}}$ R-CNN\xspace (lower is better). By adding a stronger feature pyramid from \cite{nasfpn}, we outperform Mask$^{\mathit{X}}$ R-CNN\xspace by 9.4 and 6.2 AP. This shows that ShapeMask can take advantage of large backbone model. We also observe that ShapeMask clearly outperforms the baseline class agnostic Mask R-CNN reported in \cite{hu2017learning} or our own Mask R-CNN implementation. These results provide strong evidence that ShapeMask can better generalize to categories without mask annotations.

Figure \ref{fig:vis-partial} visualizes the outputs of ShapeMask in the partially supervised setting. ShapeMask is able to capture many objects well despite not having seen any example mask of the same category during training. The mask branch was trained on VOC, tested on Non-VOC categories and vice versa. By using shape prior and instance embedding, ShapeMask is able to predict complete object-looking shapes in cases where the pixelwise prediction approaches like Mask R-CNN tend to predict broken pieces. Figure \ref{fig:vis-comparison} in Appendix A shows more results of ShapeMask vs. Mask R-CNN in the partially supervised setting.

\begin{figure}
\begin{center}
  \includegraphics[width=0.8\linewidth]{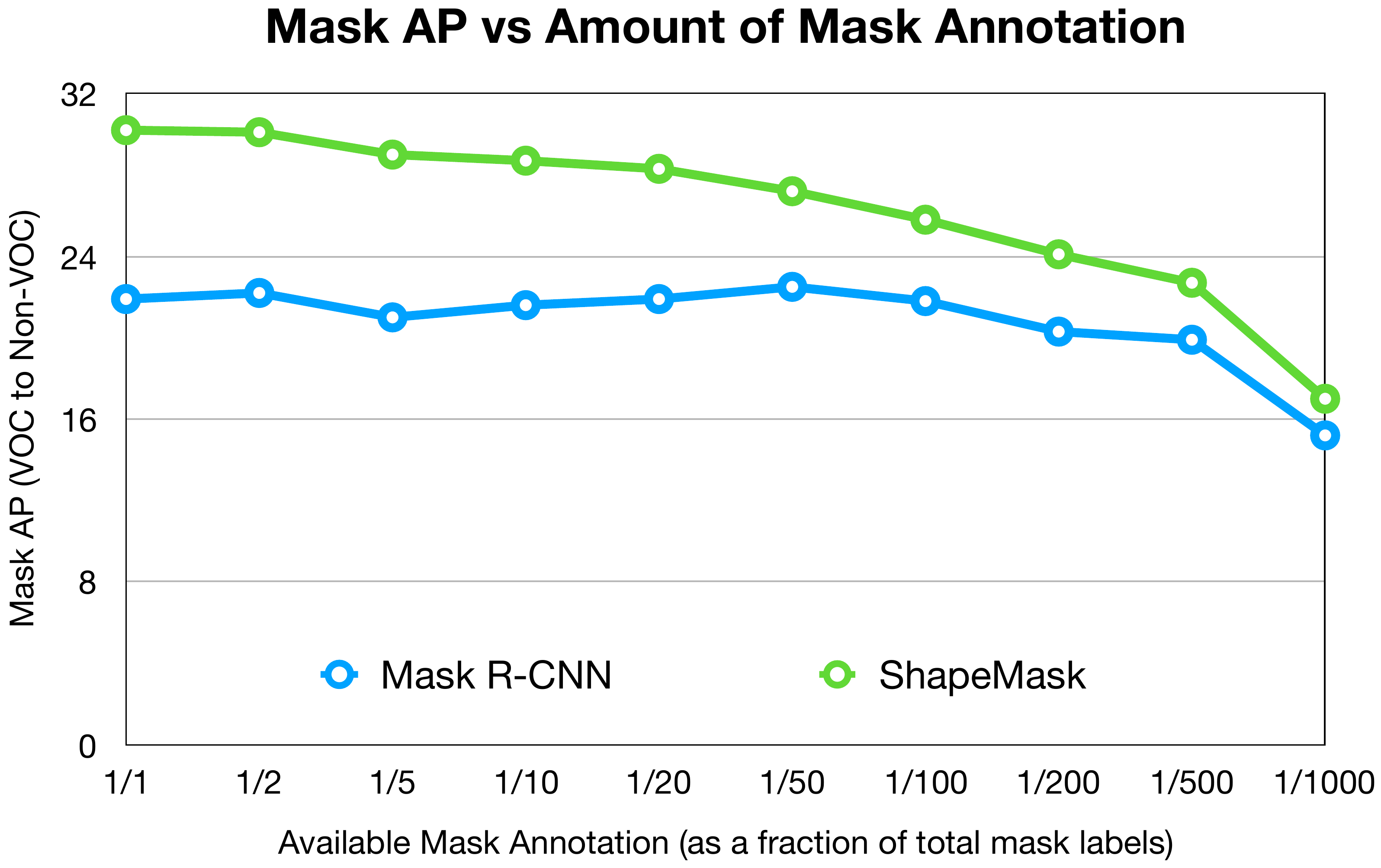}
\end{center}
\vspace{-4mm}
\caption{Generalization with less data. ShapeMask generalizes well down to 1/1000 of the training data. Using only 1/100 of the training data, ShapeMask still outperforms state-of-the-art Mask$^{\mathit{X}}$ R-CNN\xspace trained on the whole data by 2.0 AP.}
\vspace{-5mm}
\label{fig:mask_subsample}
\end{figure}

\noindent \textbf{Generalization with less data:}
\label{sec:less-annotation}
To study the generalization capabilities of ShapeMask with less training data, we train class agnostic ShapeMask and Mask R-CNN on VOC and test on Non-VOC categories using only 1/2 until 1/1000 of the data. To mimic the realistic setting of having less labeled data, we subsample the training set by their image id. Figure \ref{fig:mask_subsample} shows that ShapeMask generalizes well to unseen categories even down to 1/1000 of the training data. In fact, using just 1/100 of the training data, ShapeMask still outperforms the state-of-the-art Mask$^{\mathit{X}}$ R-CNN\xspace trained on the whole data by 2.0 AP.

\begin{figure}[t]
\includegraphics[width=1.0\linewidth]{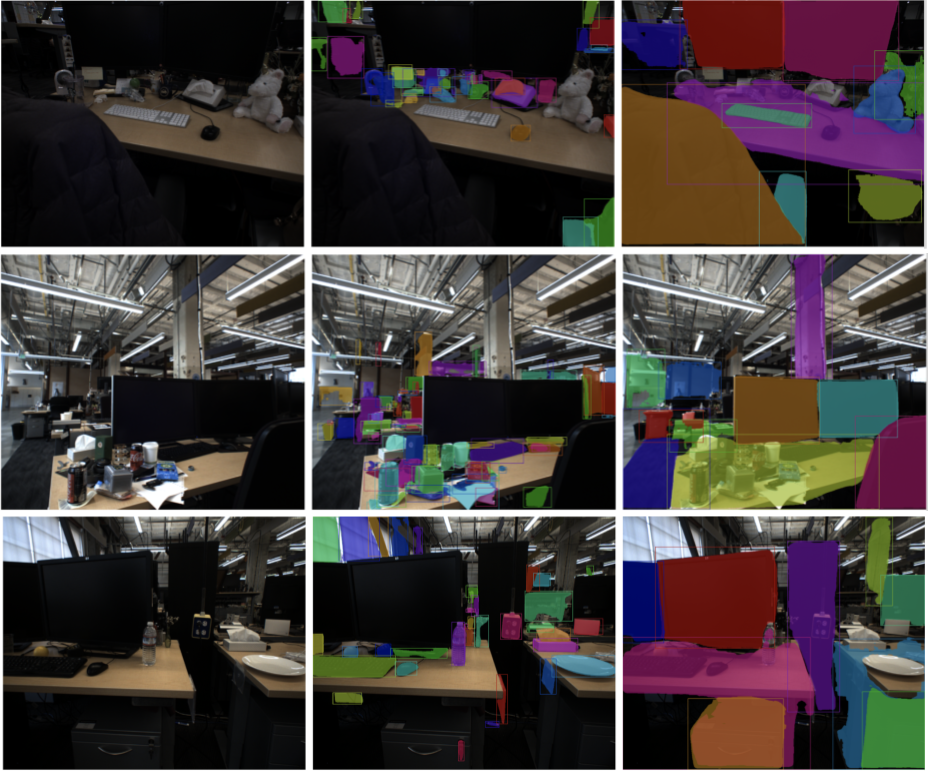}
\vspace{-4mm}
\caption{ShapeMask applied for object instance segmentation for robotics grasping. Here the ShapeMask model is trained on the COCO dataset and is not fine-tuned on data from this domain. As seen, it successfully segments the object instances, including novel objects such as a plush toy, a document, a tissue box, etc. For better visualization, smaller segmented objects are shown in the middle column and larger ones in the right column.}
\vspace{-5mm}
\label{fig:robotics}
\end{figure}

\noindent \textbf{Generalization to robotics data:}
\label{sec:robotics}
We further demonstrate the ShapeMask algorithm in an out-of-sample scenario, by testing it on object instance segmentation for robotics grasping (Figure~\ref{fig:robotics}). 
This dataset contains many objects not defined in the COCO VOCabulary, therefore serving as a good testbed to assess the generalization of ShapeMask. The dataset comes with bounding box annotations on office objects and architectural structures, but \textit{without any instance mask annotation}. The model is \textit{only trained on COCO} and not on this data. To isolate the task of instance segmentation from detection, we feed in groundtruth boxes and evaluate only on segmentation task. As seen, ShapeMask generalizes well to many categories not present in the training data.
This shows our approach is particularly useful in settings where the agent will encounter objects beyond the pixelwise annotated vocabulary.
 
\subsection{Fully Supervised Instance Segmentation}
Although the focus of ShapeMask is on generalization, this section shows that it is competitive as a general purpose instance segmentation algorithm.

\noindent \textbf{Main results:}
We compare class-specific ShapeMask to leading instance segmentation methods on COCO in Table \ref{tab:final_mask}. Following previous work ~\cite{he2017mask}, training is on COCO train2017 and testing is on test-dev2017. 

Using the same ResNet-101-FPN backbone, ShapeMask outperforms Mask R-CNN by 1.7 AP. With a stronger backbone, ShapeMask outperforms the best Mask R-CNN and MaskLab numbers by 2.9 and 2.7 AP. Since the focus of ShapeMask is to generalize to novel categories, we do not apply the techniques reported in \cite{chen2017masklab,liu2018path}, including atrous convolution, deformable crop and resize, mask refinement, adaptive feature pooling, heavier head, etc. Without any of these, ShapeMask ranks just behind PANet by 2.0 AP. Similarly, ShapeMask achieves 45.4 AP for box detection task without using the techniques reported by \cite{cai2017cascade,singh2018analysis,liu2018path} -- only second to the 47.4 AP of PANet (see Table \ref{tab:final_box} in Appendix B). 
\begin{table*}[t]
\tablestyle{3.5pt}{1.1}
\begin{tabular}{l|l|x{22}x{22}x{22}|x{22}x{22}x{22}x{22}x{44}x{44}}
 & backbone & AP & AP$_{50}$ & AP$_{75}$ & AP$_S$ &  AP$_M$ &  AP$_L$ & Training (hrs) & Inference \quad (X + Y ms) & GPU \\
\shline
 FCIS+++ \cite{li2016fully} +OHEM & ResNet-101-C5-dilate 
 & 33.6 & 54.5 & - & - & - & - & 24 & 240 & K40\\
 Mask R-CNN \cite{he2017mask} & ResNet-101-FPN 
 & 35.7 & 58.0 & 37.8 & 15.5 & 38.1 & 52.4 & 44 & 195 + 15 & P100 \\
 Detectron Mask R-CNN \cite{Detectron2018} & ResNet-101-FPN 
  & \demph{36.4} & - & - & - & - & - & 50 & 126 + 17 & P100 \\
 ShapeMask (ours)  & ResNet-101-FPN 
  & 37.4 & 58.1 & 40.0 & 16.1 & 40.1 & 53.8 & 11* & 125 + 24  & V100 \\
\hline
 Mask R-CNN \cite{he2017mask} & ResNext-101-FPN 
 & 37.1 & 60.0 & 39.4 & 16.9 & 39.9 & 53.5 & - & - & -\\
 MaskLab \cite{chen2017masklab} & Dilated ResNet-101
  & 37.3 & 59.8 & 39.6 & 19.1 & 40.5 & 50.6 & - & - & -\\
 PANet \cite{liu2018path} & ResNext-101-PANet
  & 42.0 & 65.1 & 45.7 & 22.4 & 44.7 & 58.1 & - & - & -\\
 ShapeMask (ours)  & ResNet-101-NAS-FPN \cite{nasfpn}
  & 40.0 & 61.5 & 43.0 & 18.3 & 43.0 & 57.1 & 25* & 180 + 24 & V100\\
\end{tabular}\vspace{-2mm}

\caption{ShapeMask Instance Segmentation Performance on COCO. Using the same backbone, ShapeMask outperforms Mask R-CNN by 1.7 AP. With a larger backbone, ShapeMask outperforms Mask R-CNN and MaskLab by 2.9 and 2.7 AP respectively. Compared to PANet, ShapeMask is only 2.0 AP behind without using any techniques reported in \cite{liu2018path,chen2017masklab}. This shows that ShapeMask is competitive in the fully supervised setting. Timings reported on TPUs are marked with star signs. Inference time is reported following the Detectron format: X for GPU time, Y for CPU time. All mask APs are single-model, and are reported on COCO test-dev2017 without test time augmentation except Detectron on val2017 (gray).}
\label{tab:final_mask}\vspace{-3mm}
\end{table*}

We benchmark the training and inference time with existing systems. Our training time of 11 hours on TPUs is 4x faster than all versions of Mask R-CNN \cite{he2017mask,Detectron2018} \footnote{github.com/facebookresearch/Detectron/blob/master/MODEL\_ZOO.md}. For ResNet-101 model, we report competitive inference time among leading methods, where we note that our CPU time is unoptimized and can be further reduced. Among the heavier models, ShapeMask is the only method with reported runtimes. Training finishes within 25 hours on TPUs and runs at 5 fps per 1024 x 1024 image on GPU. Table \ref{tab:ablation_params} of Appendix C further shows that by reducing the feature channels of the mask branch, we can reduce the mask branch capacity by 130x and run 6x faster there (4.6ms) with marginal performance loss. These results show that ShapeMask is among the most efficient methods. 

\begin{figure*}[t]
\begin{center}
  \includegraphics[width=1.0\textwidth]{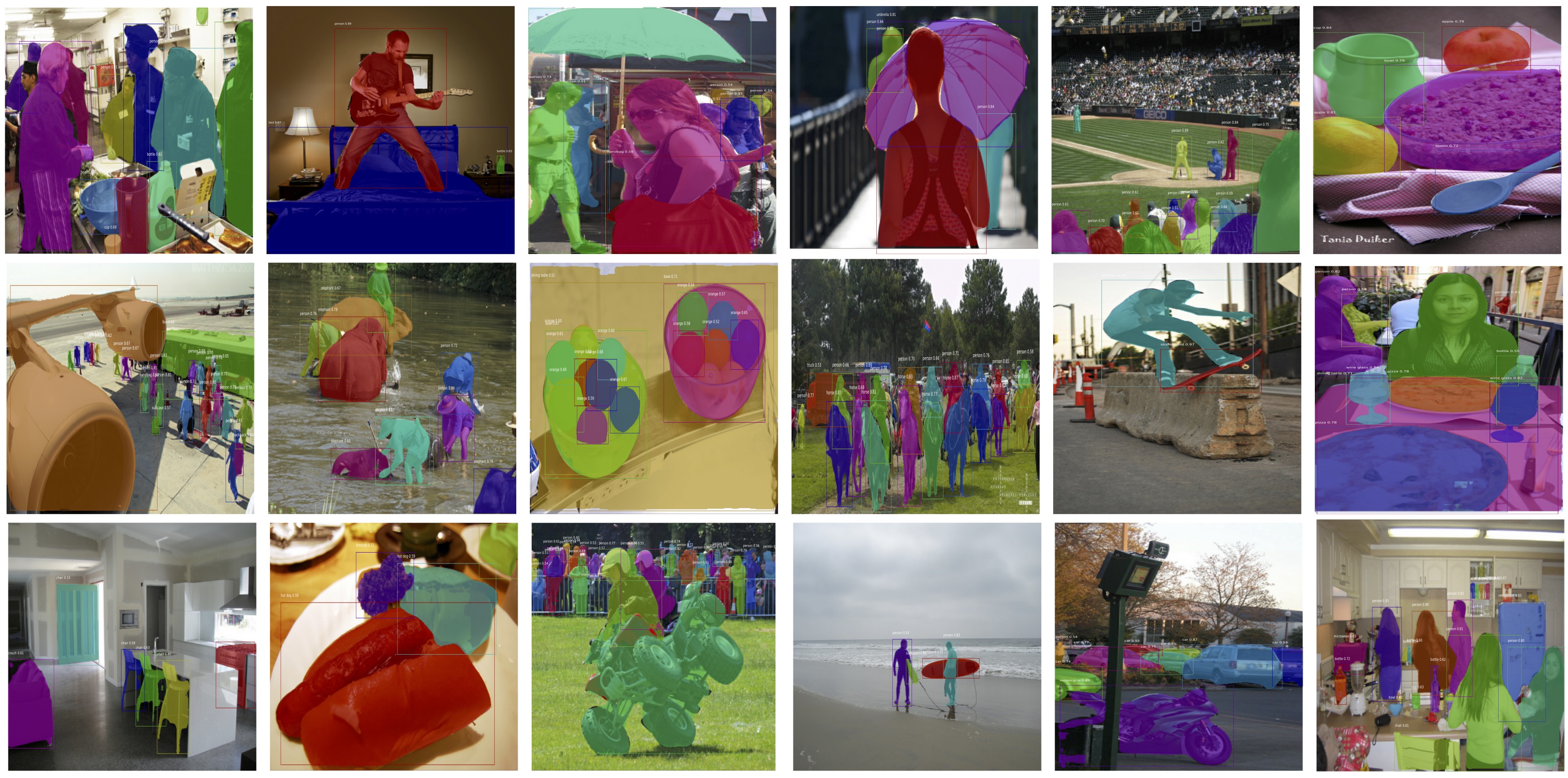}
\end{center}
\vspace{-4mm}
\caption{Visualization of results of the fully supervised ShapeMaskmodel on the COCO val2017. ShapeMask is able to obtain quality contours for large objects, handle thin structures , and deal with within-category overlaps.}
\vspace{-3mm}
\label{fig:vis-full}
\end{figure*}

Figure \ref{fig:vis-full} visualizes the outputs of the fully supervised ShapeMask. ShapeMask obtains quality contours for large objects (e.g. humans and animals) and can handle thin structures (e.g. legs of chairs, skateboard) and within-category overlaps (e.g. crowds of humans). Results are generated by class-specific ResNet-101-FPN model.

\begin{figure}[h]
\vspace{-2mm}
\begin{center}
  \includegraphics[width=1.0\linewidth]{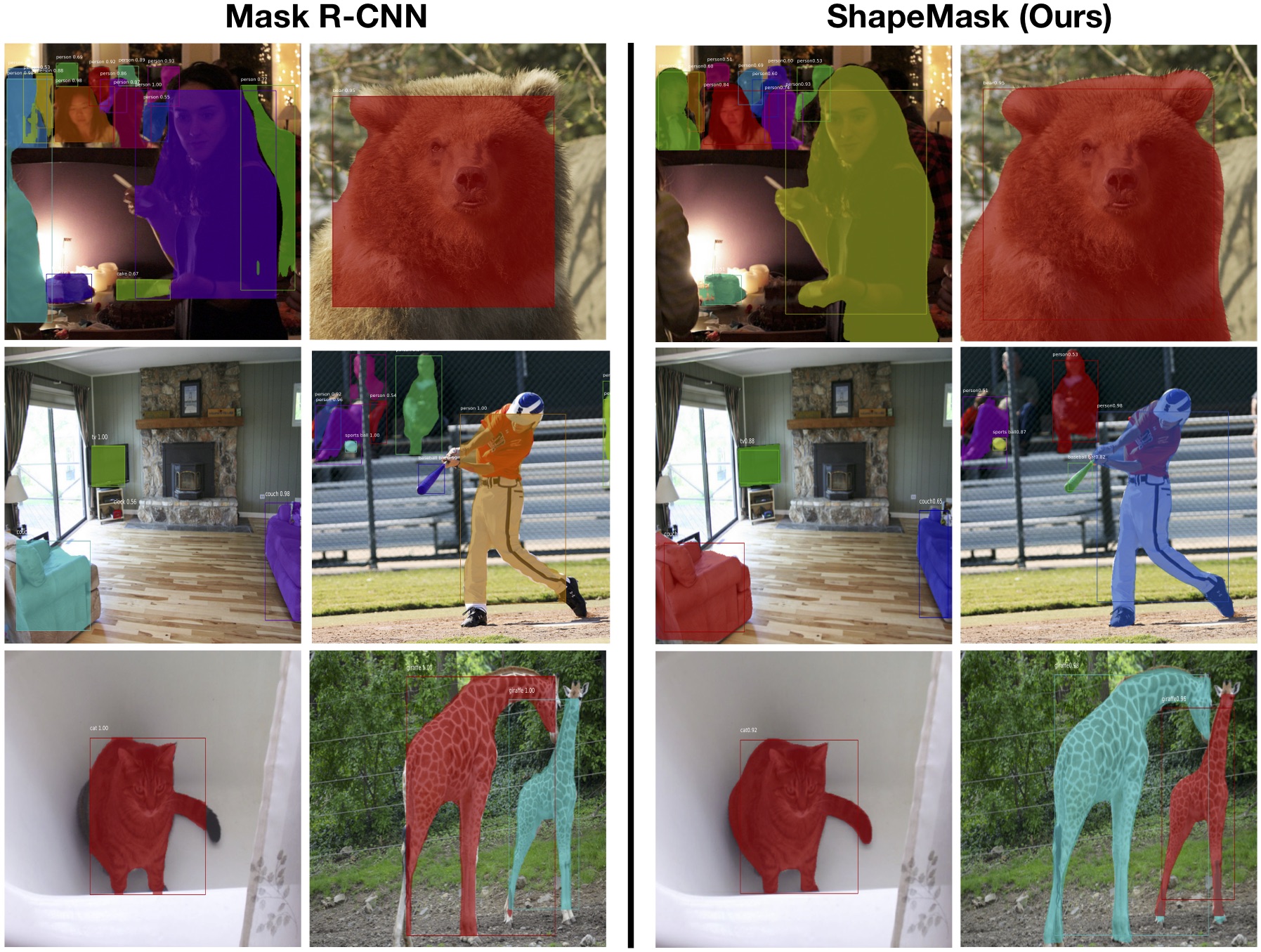}
\end{center}
\vspace{-5mm}
\caption{Analysis of Robust Segmentation. We stress-test Mask R-CNN and ShapeMask on randomly perturbed boxes (both were trained on whole boxes). Using soft detection priors, ShapeMask can handle poorly localized detections at test time while Mask R-CNN fails to do so by design of tight feature cropping.}
\vspace{-3mm}
\label{fig:vis-shrink-box}
\end{figure}

\noindent \textbf{Analysis of robust segmentation:}
With pixelwise prediction approaches such as ~\cite{he2017mask}, the quality of mask depends heavily on detection quality by design. When detections are not reliable, there exists no mechanism for the mask branch to recover. In ShapeMask, masks are not confined to come from within detection boxes. We analyze the robustness of segmentation by conducting the following experiment.

First, we perturb the box detections at inference time by downsizing the width and height independently with a random factor $x \sim U(0.75, 1.00)$, where $U$ represents uniform distribution. Downsizing avoids the complication of overlapping detections.  Figure~\ref{fig:vis-shrink-box} compares the masks produced by Mask R-CNN and ShapeMask under this perturbation. Since Mask R-CNN can only produce masks within the boxes, it is not able to handle poorly localized detections. In contrast, ShapeMask uses detection merely as soft shape priors and manage to correct those cases without being trained for it at all. In addition, Table. \ref{tab:shrink_box} quantifies the effect of downsized detections on mask quality. We see a significant drop in Mask R-CNN performance while ShapeMask remains stable. In addition, we show that training ShapeMask on downsized boxes improves the robustness.

\begin{table}[t]
\tablestyle{3.5pt}{1.1}
\begin{tabular}{l|x{44}x{44}}
 Method & No Jittering  & Jitterring \\
\shline
 Our Mask R-CNN    &36.4 & 29.0 \\
\hline
 ShapeMask (ours) &37.2 &\textbf{34.3} \\
 ShapeMask w/ jittering training (ours) &37.2  &\textbf{35.7} \\
\end{tabular}\vspace{-3mm}
\caption{Instance segmentation Mask AP with jittered detections at test time. ShapeMask is more robust than Mask R-CNN by 5.3 AP. Adding jittering during training time makes ShapeMask more robust to it (last row).}
\label{tab:shrink_box}\vspace{-5mm}
\end{table}

\subsection{Model Ablation}
To understand our system further, 
we compare the uniform box prior with the learned detection prior, and the direct mask decoding \cite{he2017mask} with the instance conditioned mask decoding.
Table. \ref{tab:ablation_partial} shows the partially supervised system ablation results on COCO val2017 using ResNet-101-FPN. Surprisingly, using either object shape prior or instance embedding greatly improves from the baseline by about 12 and 5 AP. Combining both techniques boosts the performance even further. This demonstrated the importance of the key components of our algorithm, namely shape priors and learned embeddings. Similar results are found for the fully supervised setting (see Table \ref{tab:ablation_full} in Appendix D).

\begin{table}
\tablestyle{3.5pt}{1.1}
\begin{tabular}{x{22}x{33}|x{22}x{22}x{22}|x{22}x{22}x{22}}
& & \multicolumn{3}{c|}{VOC $\rightarrow$ Non-VOC} & \multicolumn{3}{c}{Non-VOC $\rightarrow$ VOC} \\
 Shape & Embed. & AP & AP$_{50}$ & AP$_{75}$ & AP & AP$_{50}$ & AP$_{75}$\\
\shline
  & 
  & 13.7 & 28.0 & 12.0 & 24.8 & 45.6 & 23.5\\
  & \checkmark
  & 26.2 & 44.6 & 27.1 & 29.4 & 51.7 & 29.0\\
  \checkmark &
  & 26.4 & 44.9 & 27.2 & 30.6 & 53.4 & 30.4\\
  \checkmark & \checkmark
  & 30.2 & 49.3 & 31.5 & 33.3 & 56.9 & 34.3
\end{tabular}
\vspace{-3mm}
\caption{Ablation results for the partially supervised model. 
}
\label{tab:ablation_partial}\vspace{-5mm}
\end{table}

\vspace{-1mm}
\section{Conclusion}
We introduce ShapeMask that leverages shape priors and instance embeddings for better generalization to novel categories. ShapeMask significantly outperforms state-of-the-art in the cross categories setup. Moreover, it is robust against inaccurate detections, competitive in the fully supervised setting, and runs efficiently for training and inference. We believe ShapeMask is a step to further instance segmentation in the wild.

\section*{Appendix A: ShapeMask vs. Mask R-CNN}
\begin{figure*}[t]
\begin{center}
  \includegraphics[width=1.0\textwidth]{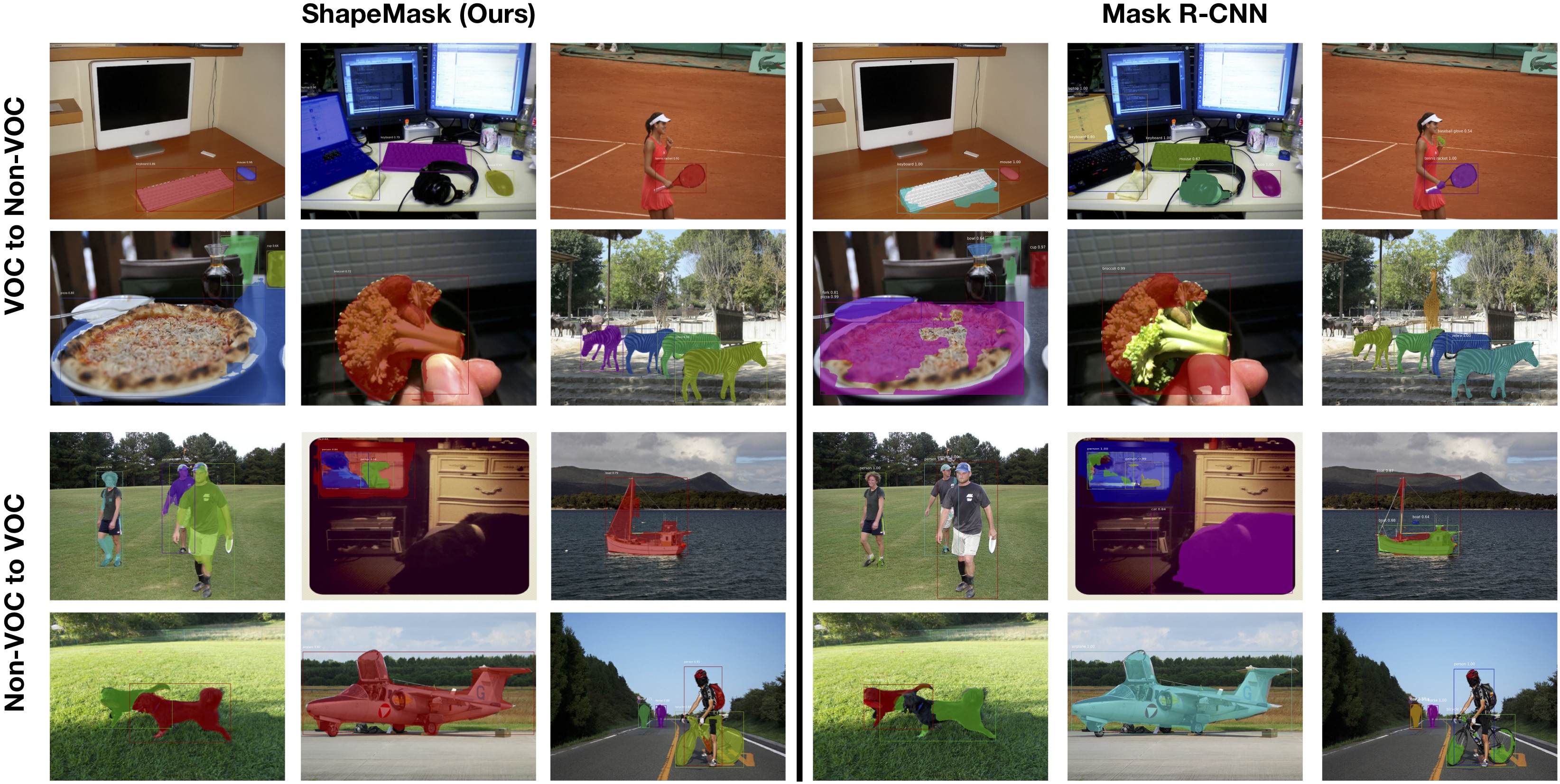}
\end{center}
\vspace{-4mm}
\caption{Random Visualization of Partially Supervised ShapeMask vs. Mask R-CNN. ShapeMask outperforms Mask R-CNN in most cases by learning shape priors and instance embeddings.}
\vspace{-3mm}
\label{fig:vis-comparison}
\end{figure*}

Figure \ref{fig:vis-comparison} shows more outputs of the partially supervised ShapeMask vs. Mask R-CNN. We randomly sample the images from the validation set and visualize VOC to Non-VOC and Non-VOC to VOC transfers. Both methods are able to segment unseen object categories, with ShapeMask clearly providing more accurate segments. Potential failure modes are exhibited by both on challenging categories such as pizza and people.
\section*{Appendix B: Object Detection}
Instance segmentation algorithms are also evaluated by their ability to provide accurate detections. Table \ref{tab:final_box} shows our comparison with leading object detectors on COCO. With ResNet-101-FPN backbone, our 42.0 AP clearly outperforms RetinaNet and Mask R-CNN, and is among the best reported approaches using the same backbone. Using ResNet-101-NAS-FPN, ShapeMask achieves 45.4 AP which is comparable to SNIP and behind PANet by 2.0 AP. Note that ShapeMask does not apply many existing detection improvement methods \cite{cai2017cascade,jiang2018acquisition,singh2018analysis}. This shows that ShapeMask can function as a competitive object detector as well. 

\begin{table*}[t]
\tablestyle{3.5pt}{1.1}
\begin{tabular}{l|l|x{22}x{22}x{22}|x{22}x{22}x{22}x{44}x{44}}
 & backbone & AP &  AP$_{50}$ & AP$_{75}$ & AP$_S$ &  AP$_M$ &  AP$_L$ \\
\shline
 Mask R-CNN \cite{he2017mask} & ResNet-101-FPN
  & 38.2 & 60.3 & 41.7 & 20.1 & 41.1 & 50.2 \\
 RetinaNet \cite{lin2018focal} & ResNet-101-FPN
  & 39.1 & 59.1 & 42.3 & 21.8 & 42.7 & 50.2 \\
 MaskLab \cite{chen2017masklab} & Dilated ResNet 101
  & 41.9 & 62.6 & 46.0 & 23.8 & 45.5 & 54.2 \\
 Cascade R-CNN \cite{cai2017cascade} & ResNet-101
  & 42.8 & 62.1 & 46.3 & 23.7 & 45.5 & 55.2 \\
 SNIP \cite{singh2018analysis} & DPN-98 
  & 45.7 & 67.3 & 51.1 & 29.3 & 48.8 & 57.1 \\
 PANet \cite{liu2018path} & Path-Agg. ResNext-101
  & 47.4 & 67.2 & 51.8 & 30.1 & 51.7 & 60.0 \\
\hline
 ShapeMask (ours) & ResNet-101-FPN
  & 42.0 & 61.2 & 45.7 & 24.3 & 45.2 & 53.1 \\
 ShapeMask (ours) & ResNet-101-NAS-FPN \cite{nasfpn}
  & 45.4 & 64.2 & 49.2 & 27.2 & 49.0 & 56.9
\end{tabular}\vspace{-2mm}
\caption{Object Detection Box AP on COCO test-dev2017. With ResNet-101-FPN backbone, ShapeMask outperforms RetinaNet and Mask R-CNN, and is among the best reported approaches using the same backbone. With a larger backbone, ShapeMask achieves comparable performance to SNIP and trails PANet by 2 point without using any techniques from \cite{cai2017cascade,jiang2018acquisition,singh2018analysis}. This shows that ShapeMask can function as a competitive object detector. All entries are single model results and without test time augmentation.}
\label{tab:final_box}\vspace{-3mm}
\end{table*}

\section*{Appendix C: Lightweight Mask Branch}
We study the performance and mask branch capacity tradeoff in Table \ref{tab:ablation_params}. All convolution and deconvolution layers in the mask branch are set to the same number of channels here. We observe that ShapeMask performance degrades minimally as mask branch capacity decreases dramatically. With 16 channels, the mask branch of ShapeMask maintains a competitive AP of 35.8, slightly better Mask R-CNN, while using 130x fewer parameters and 23x fewer FLOPs and running at 4.6ms. To our knowledge, this is the most lightweight and yet competitive mask branch design for instance segmentation.

\begin{table}
\tablestyle{3.5pt}{1.1}
\begin{tabular}{l|x{33}|x{22}|x{33}x{33}|x{22}}
 Model & \# of Chns. & AP & Params. (M) & FLOPs (M) & Time (ms) \\
\shline
 Mask R-CNN \cite{he2017mask} & 256 & 35.4 & 2.64 & 530 & - \\
\hline
 ShapeMask (ours)  & 128 & 37.0 & 1.44 & 1480 & 29.1\\
 ShapeMask (ours)  & 64 & 36.7 & 0.36 & 370 & 14.0\\
 ShapeMask (ours)  & 32 & 36.6 & 0.09 & 93 & 7.0\\
 ShapeMask (ours)  & 16 & 35.8 & 0.02 & 23 & 4.6
\end{tabular}
\vspace{-2mm}
\caption{Performance vs. mask branch model capacity. The performance decreases only slightly with a dramatic decrease in the model capacity of the mask branch. With only 16 channels, ShapeMaskmodel achieves 0.4 AP higher than Mask R-CNN with 130x fewer parameters and 23x fewer FLOPs. To our knowledge, this is the most lightweight and yet competitive mask branch design for instance segmentation.  Timing is measured on the mask branch only.}
\label{tab:ablation_params}
\end{table}

\section*{Appendix D: Ablation Studies of Fully Supervised ShapeMask}
Table \ref{tab:ablation_full} shows the fully supervised system ablation results on COCO val2017 using ResNet-101-FPN. Using either object shape prior or instance embedding improves from the baseline. Combining both techniques boosts the performance even further. This demonstrated the importance of the key components of our algorithm, namely shape priors and learned embeddings. 

\begin{table}
\tablestyle{3.5pt}{1.1}
\begin{tabular}{x{22}x{33}|x{22}x{22}x{22}}
 Shape & Embed. & AP & AP$_{50}$ & AP$_{75}$\\
\shline
  & 
  & 35.5 & 56.5 & 37.9 \\
  & \checkmark
  & 36.7 & 57.3 & 38.9 \\
  \checkmark &
  & 36.9 & 57.3 & 39.6\\
  \checkmark & \checkmark
  & 37.2 & 57.6 & 39.6
\end{tabular}
\vspace{-2mm}
\caption{Ablation results for the fully supervised model.}
\label{tab:ablation_full}\vspace{-3mm}
\end{table}

{\small
\bibliographystyle{ieee}
\bibliography{egbib}
}
\end{document}